# A Scalable Conditional Independence Test for Nonlinear, Non-Gaussian Data

Joseph D. Ramsey
Department of Philosophy
Carnegie Mellon University
Pittsburgh, PA 15221
jdramsey@andrew.cmu.edu

Abstract.

Many relations of scientific interest are nonlinear, and even in linear systems distributions are often non-Gaussian, for example in fMRI BOLD data. A class of search procedures for causal relations in high dimensional data relies on sample derived conditional independence decisions. The most common applications rely on Gaussian tests that can be systematically erroneous in nonlinear non-Gaussian cases. Recent work (Gretton et al. (2009), Tillman et al. (2009), Zhang et al. (2011)) has proposed conditional independence tests using Reproducing Kernel Hilbert Spaces (RKHS). Among these, perhaps the most efficient has been KCI (Kernel Conditional Independence, Zhang et al. (2011)), with computational requirements that grow effectively at least as $O(N^3)$, placing it out of range of large sample size analysis, and restricting its applicability to high dimensional data sets. We propose a class of $O(N^2)$ tests using conditional correlation independence (CCI) that require a few seconds on a standard workstation for tests that require tens of minutes to hours for the KCI method, depending on degree of parallelization, with similar accuracy. For accuracy on difficult nonlinear, non-Gaussian data sets, we also compare a recent test due to Harris & Drton (2013), applicable to nonlinear, non-Gaussian distributions in the Gaussian copula, as well as to partial correlation, a linear Gaussian test.

1. Introduction.

Tests of conditional independence for nonlinear or non-Gaussian data have used a reproducing kernel Hilbert space (RKHS). An excellent representative of these tests, Zhang, et al.'s KCI ("Kernel Conditional Independence"), was shown to have good accuracy for linear Gaussian data used with the well-known PC search algorithm (Spirtes et al., 2000). We find the test is accurate for nonlinear, non-Gaussian data as well. But despite the fact that the test scales better than previous such tests (Gretton et al. (2009), Tillman et al. (2009)), it does not scale well enough for general use. In general, it scales cubically with sample size, for sample size 1000 can require several seconds per test,[1] and in high dimensional problems PC and other constraint based search algorithms may call for thousands to tens of thousands of such tests.

A novel method, CCI ("Conditional Correlation Independence"), is proposed for the nonlinear, non-Gaussian case that is at least as accurate as KCI on small data sets but scales up to much larger problems. (The term "conditional correlation" is as in Baba et al., 2004. and Lawrance, 1976.) The framework of RKHS is replaced by a simpler set of mathematical ideas: The first is that X _||_ Y just in case cov(f(X), g(Y)) = 0 for all functions f and g, where X and Y are arbitrarily distributed. The second is that one only needs to test f and g in a functional basis for an appropriate space of functions. The third is that zero covariance of arbitrarily distributed X and Y can be tested using a generalized Fisher Z test of zero correlation if X and Y have finite fourth moments. The fourth idea is that X _||_ Y | Z just in case (up to a faithfulness condition) the nonparametric residuals of X regressed onto Z are independent of the nonparametric residuals of Y regressed onto Z, for models in which errors are additive. Using these ideas, a relatively simple algorithm can be devised that matches the KCI test in small sample performance and scales to large numbers of variables with sample sizes in the thousands.

Hoyer et al. (2009) and Mooij et al. (2009) also target the nonlinear, non-Gaussian case, and aim to learn both structure and orientation in same routine, though their methods do not scale up well. Under the assumption of additive noise, Hoyer et al. suggest two methods, one for pairs of variables, one for larger models, with an example given of 4 variables. Under the same assumption of additive noise, Mooij suggest a more streamlined method for larger models, with an example provided using 7 variables. The pairwise method of Hoyer et al. considers an adjacency X---Y where, say, Y = f(X) + e, where e _||_ X. In general, as they show, except for a few special cases, when f is nonlinear there is no backwards model X = g(Y) + e', where e' _||_ Y. Thus a simple test suffices. Calculate the residuals $r_X$ of regressing X nonlinearly onto Y and the residuals $r_Y$ of regressing $r_Y$ nonlinearly onto X. If $r_X$ is independent of Y but $r_Y$ is not independent of X, orient Y-->X; if the reverse is true, orient X-->Y; if both $r_X$ is independent of Y and $r_Y$ is independent of X, do not orient the edge. If $r_X$ is dependent on $r_Y$ and $r_Y$ is dependent on $r_X$, the graph is, as they say, more complicated. The method for larger numbers of variables (they suggest up to 7) is to consider all possible DAGs over the variables, calculate for each DAG the residual of each variable conditional on its parents, and accept models for which all of these residuals are independent. DAGs for which an accepted submodel exists are not returned. Since this algorithm does not scale well to many variables, it is not further considered as a method for learning structure. The Mooij et al. approach first sorts the variables high to low by their residuals regressed nonlinearly

---

[1] On a Macbook Air, 2Ghz, Matlab R2012b, using one processor.
[2] Note that it is plausible to use either the Hermite polynomials or the Hermite functions directly in

onto all of the other variables, producing a causal order, and then uses independence of residuals tests to remove unwanted edges. The first step of this algorithm, however, requires that each variable be regressed non-parametrically on all other variables in the graph, a step that does not scale well statistically. We do not for this reason further consider this algorithm for learning structure; the PC-Stable algorithm in this paper, using the CCI independence test, is intended to scale to hundreds of variables for sparse graphs. Variations of the Hoyer et al. and Mooij et al. methods are useful, however, for learning orientations for large graphs once structure is known, a topic that we will purposefully set beyond the scope of the current essay, since it deserves further elaboration.

The rest of the paper will proceed as follows. In Section 2 each of the conditions just mentioned is considered and given some justification. In Section 3 the CCI test and an argument for its correctness are given. Section 4 briefly reviews the structure of the Zhang et al. (2011) KCI test and the Harris & Drton (2013) test. Section 5 briefly reviews the PC (Spirtes et al., 2000) and PC-Stable (Colombo and Maathuis, 2013) algorithms. Section 6 compares partial correlation (Fisher Z), the Harris & Drton test, CCI and KCI on a problem approximating one proposed by Zhang et al., for linear Gaussian data. Section 7 compares these same four independence tests (in the context of PC) on various nonlinear, non-Gaussian models and gives some results of larger models with Fisher Z, Harris & Drton, and CCI. Section 8 gives a brief discussion.

2. Motivation

Zhang et al. (2011) make the following comment regarding their use of Reproducing Kernel Hilbert Spaces (RKHS): "Alternatively, Daudin (1980) gives the characterization of CI by *explicitly enforcing the uncorrelatedness of functions in suitable spaces*, which may be considered more appealing" (p. 3, sic). They point out that the relevant functions can be constructed by nonlinear regression, producing (in the large sample limit) functions of the form $r(X|Z) = X - E(X|Z)$. This idea can be followed in a somewhat different way than the authors of this paper do, avoiding appeal to RKHS and thus reducing the complexity of the procedure.

To see that $X \perp\!\!\!\perp Y$ for X and Y arbitrarily distributed, just in case $cov(f(X), g(Y)) = 0$ for all functions f and g, and that one only needs to appeal to a set of basis functions for the square summable space of transformations of a variable, we cite Daudin (1980), where
$\Phi^2_{X_1X_2} = \int f_2(x_1, x_2) \, dP_{X_1}(x_1) \, dP_{X_2}(x_2) - 1$ is a measure of association between $X_1$ and $X_2$,
$L^2_{X_1} = \{r(X_1) \text{ s.t. } E(r^2) < +\infty\}$ and $L^2_{X_2} = \{s(X_2) \text{ s.t. } E(s^2) < +\infty\}$:

> Let $\{\varphi_i\}$ for $i \in I$, where I is an index set, be a complete orthonormal system of $L^2_{X_1}$ and $\{\psi_j\}$ for $j \in J$ be a complete orthonormal system of $L^2_{X_2}$ with $\varphi_0 = \psi_0 = 1$. The term 'orthonormal' refers to the covariance inner-product and is then equivalent to 'uncorrelated'. Lancaster (1969, pp. 89-93) proved the following results:
> (i) $\Phi^2_{X_1X_2} = 0$ is equivalent to $X_1$ and $X_2$ being independent vectors;
> …
> (iii) $\Phi^2_{X_1X_2} = \Sigma_{i,j \neq 0} \{E(\varphi_i\psi_j)\}^2$. (Daudin, p. 582)

Note that (i) is a criterion for $X_1$ and $X_2$ being independent--namely, that the $\Phi^2$ association measure comes out to be zero. From (iii), $\Phi^2_{X_1X_2}$ is the sum of expected values of products of specific transformations of X and of Y--namely, functions that form an orthonormal basis for the set of square summable transformations over X and Y. By Daudin and Theorem 1, below, cov(x, y) is nonzero just in case E(x, y) is nonzero, so it suffices to check zero covariance. Following on Daudin's comment about orthonormality, the set of (orthogonal) Hermite polynomials will do, defined recursively as $H_n(x) = 2x\, H_{n-1}(x) - 2(n-1)H_{n-2}(x)$, with $H_0(x) = 1$ and $H_1(x) = 2x$ (Bunck, 2009). Note that $cov(H_m(x), H_n(x)) \neq 0$ (excluding unity, as Daudin does) just in case $cov(P_{m'}(x), P_{n'}(x)) \neq 0$ for some m', n', for $P = \{P_i(x) = x^i$ s.t. i = 1, 2, 3, ...\}, up to faithfulness (exact cancellation of terms); therefore, the latter set P will suffice for our purposes. For instance, if $cov(x, y^2) \neq 0$, then for some Hermite polynomials $H_f(x)$ and $H_g(x)$, $cov(H_f(x), H_g(x)) \neq 0$, specifically $cov(2x, 4y^2 - 2) = 8\, cov(x, y\verb|^|2) - 4\, cov(2x, 1)$. From the latter we conclude that $\Phi^2_{X1X2} \neq 0$, and hence X1 and X2 are dependent. If no such combination of functions from P of X1 and X2, respectively, yield covariances different from zero, then up to faithfulness, X1 is independent of X2.[2]

One can see how Daudin's condition (iii) works in practice by considering the standard example of a nonlinear distribution $y = x^2 + e$, where $x \sim U(-2,2)$ and $e \sim U(-.5, .5)$; see Figure 1. Notably, in this plot, $cov(x, y) = 0$, despite the fact that y clearly depends on x; one can see this by noting that the ordinary least squares regression line through the points has approximately zero slope. Nevertheless, as the value of x increases from -2 to 2, the mean of y changes. If one squares x, one gets the plot in Figure 2; in this case, $cov(x^2, y)$ shows a linear dependence; $cov(x^2, y) = 1$, and the ordinary least squares regression line through these points has positive slope. According to Daudin's condition (i), this fact is sufficient to show that x and y are dependent; if x and y were independent, no functional transformation of x and/or y could yield a nonzero trend. Our algorithm uses this intuition rather directly to estimate conditional independence.

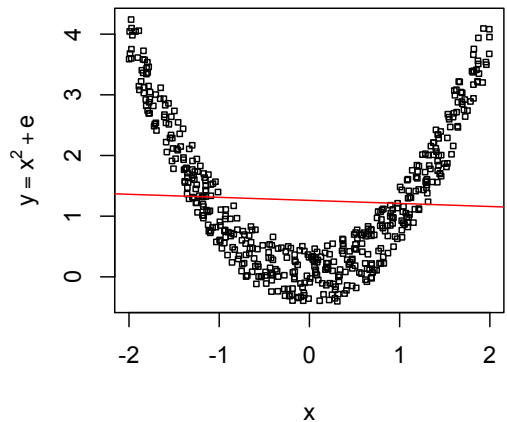

Figure 1. Scatterplot of $y = x^2 + e$ against x, as described in the text.

---

[2] Note that it is plausible to use either the Hermite polynomials or the Hermite functions directly in the algorithm below; the set P offers better performance, and so we prefer it. Another plausible basis to use is a sine/cosine basis (Lanczos, 1988).

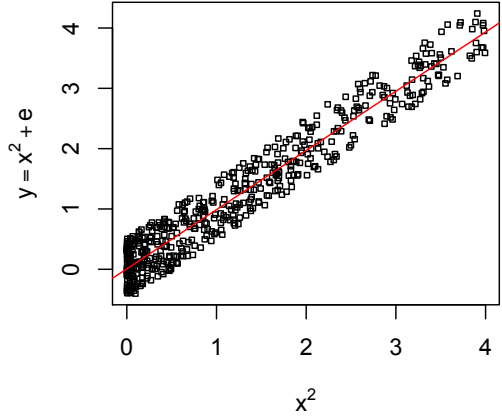

Figure 2. Scatterplot of y=x² +e against x².

The third idea is to use a generalized Fisher Z model to test for zero correlation. One of course does not test directly whether cov(X, Y) = 0; one instead tests whether corr(X, Y) = 0, since correlation is appropriately scaled. With X and Y Normally distributed, the usual test for corr(X, Y) = 0 calculates the correlation r of X and Y and then applies the Fisher Z transform, $f(r) = 1/2 \ln((1+r)/(1-r))$; in the large sample limit $N^{1/2} f(r) \rightarrow N(0,1)$, so one can obtain a p value and compare it to an alpha level to produce a judgment of zero correlation. When X and Y are arbitrarily distributed, but with finite fourth moments, a similar test has been suggested by Hawkins (1989) based on U-statistics. One calculates a correlation between X and Y, as above, and calculates its Fisher Z transform f(r), but refers f(r) to a Normal distribution with a different variance $\tau^2$. To calculate $\tau^2$, one calculates $\tau^2 = \Sigma(std(X)^2 std(Y)^2)$, where std(X) is the standardization of X and std(Y) is the standardization of Y. This yields a p value in the usual way, which can be compared for large N to an alpha level to produce a judgment of zero correlation. The limiting correctness of this formulation is given in Hawkins.

The fourth idea is that one can test whether X $\_||\_$ Y | Z by testing whether X - E(X|Z) $\_||\_$ Y - E(Y|Z), under a faithfulness condition--or, in other words, in the limit, whether the nonparametric residuals of X regressed onto Z are independent of the nonparametric residuals of Y regressed onto Z, for additive models. This is common knowledge, cited by Zhang et al., but it is useful to understand why it is the case in this context. By definition, X $\_||\_$ Y | Z just in case X $\_||\_$ Y | Z = z for any z. Consider values Z = z1 and Z = z2. Since X $\_||\_$ Y | Z = z1 and X $\_||\_$ Y | Z = z2, it follows that cov(f(X), g(Y)) = 0 for each of Z = z1, and Z = z2, for any transformations f, g. Let x = f(X), y = g(Y) for Z = z1 and z = f(X), w = g(Y) for Z = z2; then N cov(x, y) = $\Sigma$(xy - E(xy)) = 0 and N cov(z, w) = $\Sigma$(zw - E(zw)) = 0 imply that $\Sigma$((xy + zw) - E(xy + zw)) = $\Sigma$(xy - E(xy)) + $\Sigma$(zw - E(zw)) = 0 = 2N cov(f(X), g(Y)) for Z = {z1 or z2}. So X $\_||\_$ Y | Z = z1 and X $\_||\_$ Y | Z = z2 imply that X $\_||\_$ Y | Z = {z1 or z2}. This simple argument can be extended to the population. In order for the converse to hold, however, one needs a faithfulness condition to the effect that a zero sum in the aggregate is not achieved by accidental canceling of effects in the population, since it's possible that $\Sigma$((xy + zw) - E(xy + zw)) = 0 for the aggregate even if $\Sigma$(xy - E(xy)) = 1 and $\Sigma$(zw - E(zw)) = -1. We make this assumption.

For finite samples, one cannot actually know the distribution of X or of Y conditional on a particular value z of Z; one has single measurements for particular values of Z. This issue can be addressed by applying a kernel to the data, using a kernel width that approaches zero as sample size increases to infinity. The residual of X regressed onto Z of a particular datum is the difference between the value of X for that datum and the predicted value of X for that datum, calculated as the weighted average (the kernel providing the weights) of points whose Z coordinates are close in distance to the Z coordinates for the given datum. Choice of kernel function can require some thought for particular cases; for our purposes, to make the large sample convergence claim easy, a Uniform kernel is used. For kernel widths, we use widths of a form recommended by Bowman and Azzalini (1997), using a scaling of Mean Absolute Deviation (MAD) for each variable guaranteed to converge to zero in the large sample limit, h = 1.4826 * MAD * ((4/3) / N) ^ 0.2, converges to zero as N becomes large. To calculate the MAD of a variable, first calculate the median of the variable, then calculate the absolute difference of each point from its median, then calculate the median of these absolute differences. Scaled appropriately, MAD may be used as a substitute for standard deviation for Normal distributions and for non-Gaussian data as well. Also, when the regressing set Z has more than one variable, it is necessary to combine multiple kernel widths. We do this by taking the maximum of the widths for each dimension and multiplying by the square root of M.

Since on our approach a number of p values are calculated in series for each independence test (one for each combination of basis functions), a False Discovery Rate calculation is applied to determine an optimal cutoff for judging dependence (Benjamini and Hochberg, 1995).

Next, we give the algorithm, which implements these ideas.

3. CCI Algorithm

Pseudocode for the CCI algorithm is given in Box 1 and Box 2. Box 1 shows the procedure for calculating residuals of X nonparametrically regressed onto Z using a kernel regression. Box 2 gives code for the main section of the CCI algorithm. First, residuals are calculated for X regressed onto **Z** and Y regressed onto **Z**. A list of p values is started; each time a p value is calculated, it is added to this list for purposes of calculating an FDR cutoff. As discussed earlier, a truncation of the set $\{x^1, x^2, x^3, ...\}$ may be used. Next, for each combination of basis transformations f(X) and g(Y), a correlation is calculated, and its Fisher Z transform z is computed. As discussed, sqrt(N) * z approaches $N(0, \tau^2)$ in the limit of large sample; the limiting p value is calculated for z in this distribution, after calculating $\tau^2$. Finally, p is added to the list of p values. Once all p values have been collected up from this process, they are sorted low to high and an FDR cutoff computed. If all p values are greater than this cutoff, a judgment of independence is returned; if any of them fall below the cutoff, a judgment of dependence is returned.

Some notes on performance: (a) Checking all combinations of functions from the truncated basis can be time consuming, which is one reason to keep the list of functions reasonably short. (b) Calculating p values for all combinations of functions and then applying FDR is considerably slower than returning a judgment of dependence the first time a p value greater than alpha is observed, though the latter leads to lower accuracy of the test.

*Theorem 1*. Let F' = $\{f'_1, f'_2, f'_3, ...\}$ be an orthogonal basis for the set of square summable functions, excluding unity, and let X, Y, and set **Z** be continuous variables. Assume E(X | {Y} ∪

Box 1. Pseudocode for the nonparametric regression residual algorithm used in CCI. X and set **Z** consist of data vectors for distinct variables; X[i] is the ith data point for X; **Z**[i] is the ith row in the data for variables **Z**. D is a rectangular, continuous data set.

```
Procedure Residuals(X, Z, D)
    1. If |Z| = 0 return X
    2. Else
           a. residuals ← <>
           b. For i = 1 to |X|
                   i.   sum ← 0
                   ii.  weight ← 0
                   iii. For j = 1 to |X|
                            1. d ← distance(Z[i], Z[j])
                            2. k ← kernel(d)
                            3. sum ← sum + k * X[j]
                            4. weight ← weight + k
                   iv.  residuals[i] ← X[i] - sum / weight
           c. Return residuals
```

Z} and E(Y | {X} ∪ Z} are continuous. Then CCI returns correct judgments of conditional independence almost surely, in the limit of large sample, as alpha goes to zero.

*Proof Sketch.* Correctness is being approached in several dimensions. First, residuals calculated for an interval about Z = z0 approach residuals rX = X - E(X|Z) and rY = Y - E(Y|Z) in the large sample limit almost surely. Second the Fisher Z test converges to the truth in the large sample limit, as Hawkins showed. Third the truncated bases approach the entire basis, unity excluded, as f increases without bound. Thus, in the limit of large sample, as F approaches a complete basis, excluding unity, and as alpha goes to 0, if $r_X$ are the residuals for X given Z and $r_Y$ are the residuals for Y given Z, then X _||_ Y | **Z** implies that cov($f_m(r_X)$, $f_n(r_Y)$) = 0 for all m, n. Thus, if cov($f_m(r_X)$, $f_n(r_Y)$) ≠ 0 for some m, n, it must not be the case that X _||_ Y | **Z**. That X _||_ Y | Z implies that for some m, n, cov($f_m(r_X)$, $f_n(r_Y)$) ≠ 0 can be established by construction under the continuity assumptions.[3]

Note that CCI is quadratic both in sample size and in the number of basis functions used.

---

[3] Such a construction might intuitively be begun as follows. WLOG assume that $r_Y$ depends on $r_X$ and consider the distribution of <$r_X$, $r_Y$>. As x' increases from a lower bound to an upper bound, Y - E(Y|Z,X = x') will change continuously. Divide the interval of variation for $r_X$ into segments where Y - E(Y|Z,X = $x_0$) is increasing, segments where it is decreasing, and segments where it is constant. (Since there is dependence, assuming continuity of expectation, at least some segments must be increasing or decreasing.) Without violating continuity, for increasing segments, say x' in ($x_1$, $x_2$), set f(x) to a line segment from ($x_1$, 0) to ($x_2$, 1), for decreasing segments x' in ($x_3$, $x_4$), set f(x) to a line segment from ($x_3$, 1) to ($x_4$, 0), and for constant segments x' in ($x_5$, $x_6$), set f(x) to a constant function. Then cov(f($r_X$), $r_Y$) > 0. The same would be true if f(x) were smoothed appropriately to make it everywhere differentiable, yielding $f_2(x)$, in which case $f_2(x)$ could be expressed using a complete orthogonal basis as discussed.

Box 2. Pseudocode for the CCI algorithm. D is a rectangular, continuous data set; X, Y, and set **Z** consist of data for distinct variables; alpha is a significance level, by default 0.05; F is a list of functions, by default $\{x^1, x^2, x^3, ..., x^7\}$.

```
Procedure Independent(X, Y, alpha, F, D)
    1.  pList ← <>
    2.  For each combination of functions <f, g> drawn from F
            a.  r ← cov(f(X), g(Y)) / sqrt(var(f(X)) var(g(X)))
            b.  z ← 0.5 * ln((1+r)/(1-r))
            c.  X' ← (f(X) - mean(f(X))) / std(f(X))
            d.  Y' ← (g(Y) - mean(g(Y))) / std(g(Y))
            e.  τ² ← E(X'² Y'²)
            f.  p ← 2 * (1 - abs(NormalCDF(0, τ²)(sqrt(N) * z)))
            g.  Add p to pList
    3.  Apply FDR to pList with significance alpha to produce a cutoff c
    4.  If all p values in pList are greater than c, return independent
    5.  else return dependent

Procedure CCI(X, Y, Z, alpha, F, D)
    1.  r_X ← Residuals(X, Z, D)
    2.  r_Y ← Residuals(Y, Z, D)
    3.  Return Independent(r_X, r_Y, alpha, F, D)
```

4. Review of KCI and RPC

Theory for the KCI test is explained in Zhang et al. (2011); here we simply rehearse the theorems on which the test is based, to indicate the nature of the test and to reflect on its complexity. There is a separate test for X _||_ Y and for X _||_ Y | Z. The first test is given by Theorem 4 from Zhang et al., which we give here as Theorem 2:

*Theorem 2 [Independence test]*. Under the null hypothesis that X and Y are statistically independent, the statistic $T_{UI} = 1/n \, \text{Tr}(K_X K_Y)$ has the same asymptotic distribution as
$T'_{UI} = 1/n^2 \sum_{i,j=1}^{N} \{\lambda_{x,k} \lambda_{y,j} z^2_{ij}\}$.

In the first formula $K_X$ and $K_Y$ are kernel matrices for X and for Y, respectively, under a positive definite kernel, typically Gaussian. In the second formula, $\lambda_{x,k}$ and $\lambda_{y,j}$ are eigenvalues of $K_X$ and of $K_Y$, respectively, where $\lambda_{x,1} \geq ... \geq \lambda_{x,n}$ and $\lambda_{u,1} \geq ... \geq \lambda_{y,n}$, and $z^2_{ij}$ are i.i.d. $\chi^2_1$ variates. Up to issues with numerical estimation, this requires that two kernel matrices be computed and multiplied together, their eigenvalues be computed and sorted, with a null distribution calculated for the right hand side, and some simple calculations done with this information, effectively an $O(N^3)$ operation.

The conditional test is more involved; this is given in Proposition 5 of Zhang et al. (2011), which we rehearse as Theorem 3:

*Theorem 3 [Conditional Independence]*. Under the null hypothesis that X and Y are conditionally independent given Z, we have that the statistic $T_{CI} = 1/n \, \text{Tr}(K_{X|Z} K_{Y|Z})$ has the same asymptotic distribution as $T'_{CI} = 1/n \sum_1^{N^2}(\lambda'_k z^2_k)$, where $\lambda'_k$ are the eigenvalues of

$w^* w^{*T}$ and $w^* = [w^*_1,...,w^*_n]$, with the vector $w^*_t$ obtained by stacking $M^*_t = [\psi_{X|Z,1}(x^*),...,\psi_{X|Z,n}(x^*)]^T \cdot [\varphi_{X|Z,1}(x^*),..., \varphi_{X|Z,n}(x^*)]$.

For details of this theorem, please see Zhang et al. (2011). This involves calculating two conditional kernel matrices, doing a matrix multiplication, doing some linear operations, then forming the given stacked matrix and using it to calculate a null distribution, altogether $O(N^3)$.

Theoretically, the main advantage of KCI and other kernel-based independence tests is that they are correct in the large sample limit without further approximation, whereas CCI requires an finite (and therefore theoretically approximate) basis of functions.

The RPC (Rank PC) algorithm is the PC algorithm using a revised partial correlation test due to Harris & Drton (2013). The test calculates a matrix $\Psi$ over the variables for an independence question $X \perp\!\!\!\perp Y | S$, where $\Psi_{uv}$ is the Spearman's rank correlation (or alternatively Kendall's tau) of variables u and v in the set $\{X, Y\} \cup S$, and then uses this correlation to calculate partial correlations as

$\rho_{uv|S} = -\Psi_{uv}^{-1} / \sqrt{-\Psi_{uu}^{-1} \cdot -\Psi_{vv}^{-1}}$

(Harris & Drton, 2012). Since the calculation of correlations is dependent only on the order of data pairs in the ranking, the Harris & Drton test generalizes the usual partial correlation to a broader range of distributions, called the Gaussian copula, which consists of distributions that can be reached from distributions in a Gaussian random field by applying monotone increasing functions to each Gaussian in the field. We use Kendall's tau to implement this independence test and use it in the context of the adjacency search in the PC algorithm, implementing RPC. The complexity of the test is limited by the complexity of the Kendall's tau calculation; the authors give reference to an $O(n \log n)$ algorithm for this. For further details, see Harris & Drton (2013).

5. PC and PC-Stable

The PC algorithm (Spirtes et al., 2000) finds equivalence class of directed acyclic graphs (DAGs) from conditional independence facts supplied via an oracle. The basic graph theory is as follows. A graph G over a set of nodes V consists of a list of edges; for X and Y in V, X->Y is a directed edge, X---Y an undirected edge, and X<->Y a bidirected edge. Edges connecting successive nodes form *paths*; paths can be of various sorts. X->Y<-Z is a *collider*, and X->Y->Z, X<-Y<-Z, and X<-Y->Z are *noncolliders*. A chain of edges connecting nodes <X1,...,Xn> is called a *path* from X1 to Xn; a path of the form X1->...->Xn is called a directed path from X1 to Xn. An *cyclic path* is a directed path in which some variable appears twice. A *directed acyclic graph (DAG)* is a graph in which there are only directed edges and no cyclic paths. We say that X is *d-separated from* Y conditional on a set **Z** of variables in V just in case along every path from X to Y, every collider is neither in **Z** nor has a descendant in **Z**. A *pattern* (or CPDAG or essential graph) is a graph containing only directed and undirected edges representing an equivalence class of DAGs, such that a directed edge appears in the pattern just in case it appears in every member of the equivalence class, and an undirected edge X---Y appears in the pattern just in case X is adjacent to Y in every member of the equivalence class.

The PC algorithm takes as input conditional independence facts of the sort supplied by Fisher Z, Harris & Drton, KCI, or CCI and produces an estimate of the pattern of the DAG of the generating model for the data analyzed. The algorithm consists of two phases, an adjacency phase and an orientation phase. The adjacency phase begins with a complete undirected graph, then removes edges that are independent conditional on the empty set, then of edges X---Y that remain, removes those that are independent conditional on one other variable Z adjacent to X or to Y--thus, X _||_ Y | Z--then similarly conditioning on two adjacent variables, and so on, until no more edges can be removed from the graph. The orientation phase orients *unshielded colliders* X->Y<-Z with X not adjacent to Z by locating the conditional independence fact used to remove the edge X---Z in the adjacency phase, say X _||_ Y | **Z**, and asking whether the variable Y is in that set. If it is not, the path is oriented as a collider; otherwise, it is left unoriented. Once all colliders are oriented, a rule set is applied to orient further noncollider and collider paths that are implied, avoiding orientations that would introduce new unshielded colliders to the graph (since these should all have been oriented already). The result of the PC algorithm is a pattern, which can then be compared to the pattern for the DAG of the true generating model.

We also make use of the PC-Stable algorithm (Colombo and Maathuis, 2013), which modifies the PC algorithm in one important step to reduce the problem of order-dependence of the output pattern. Calling a *depth* the procedure of the PC adjacency algorithm conditioning a certain number D of variables, depth 0 is performed, then going into depth 1, a copy of the graph is made going in, and all judgments of adjacency in that depth are made with respect that that copied graph, then similarly for depth 2, and so on. The effect of this for our purpose is that the model can usefully be parallelized, since judgments of removing an edge in one part of the graph at a given depth are independent of judgments of removing an edge in another part of the graph at the same depth.

We use the PC and PC-Stable algorithms primarily to estimate adjacencies in Sections 6 and 7, mainly because the orientation information PC gives using these independence tests is not as accurate as one would like.[4] Nonlinear, non-Gaussian variants of the algorithms in Hoyer et al. (2009) and Mooij et al., (2009) may be used to accurately orient in such cases, though these will be taken up (as mentioned above) in a separate essay.

6. Comparison on Linear, Gaussian Data

The CCI test is itself quite simple, with straightforward justification; some effort needs to put therefore into showing its empirical adequacy. One comparison is to show how it performs on linear, Gaussian data. Four conditional independence tests suggest themselves, as suggested earlier: a partial correlation algorithm (Fisher Z), Harris & Drton (RPC), KCI, and CCI. We do a test of the form suggested by Zhang et al. (2011), Section 4.2.1. All tests are performed using the implementation of PC in the TETRAD freeware (Ramsey et al., 2013). The benchmark implementation of KCI is written in Matlab (MATLAB and Statistics Toolbox Release 2012a); it is referenced in the TETRAD freeware directly using the Matlab Builder JA. Harris & Drton, CCI, and Fisher Z are all implemented directly in Java, so that all algorithms are being compared using the same implementation of PC. We compare all 4 independence tests of 100 random data sets each for sample sizes 100, 250, 400, 550, and 700. Data was generated in the TETRAD freeware as follows. We create graphs of four ordered variables X1, X2, X3, and X4, by adding edges from previous variables to later variables in the list with probability 0.5. We then form linear structural equation models using these graphs, with Gaussian disturbance terms with mean zero and standard deviations drawn from U(0.1, 0.6) and coefficients drawn from U(-2, 2). Finally, we generate i.i.d., samples of the specified size.[5] The PC algorithm (TETRAD freeware) is then run on each data set, producing an estimated pattern. Also, the true pattern is calculated from the generating model of the data set on which PC is run. Finally adjacency precisions and recalls are calculated comparing the estimated pattern to the true pattern. For adjacency precision, the number of correct adjacencies in the pattern is calculated (TP), and the number of false positive adjacencies, that is adjacencies that appear in the estimated pattern but not in the true pattern (FP), and precision is calculated as TP / (TP + FP). For recall, the number of false negative adjacencies is calculated, that is, adjacencies that appear in the true pattern but not in the estimated pattern (FN), and recall is calculated as TP / (TP + FN). Results are shown in Figure 3. An alpha value of 0.01 was used, as suggested by Zhang et al. (2011)[6]. For CCI, a polynomial basis excluding unity was used, $\{x^1, x^2, x^3, x^4, x^5, x^6, x^7\}$, as described above.

---

[4] Precision of orientation information for the linear, Gaussian case is fairly good, in the 0.7-0.8 range, but similar precisions for the nonlinear, non-Gaussian cases are much worse, to the point of being useless in most cases. This is possibly due to the problem of *weak transitivity* that arises with respect to generalized structural equation models (SEMs) when connection functions are contracting, and coefficients and errors all in the range (-1, 1). In such cases, the influence of variables along transitive paths can be greatly diminished, leading to an overabundance of collider orientations in the PC orientation phase.

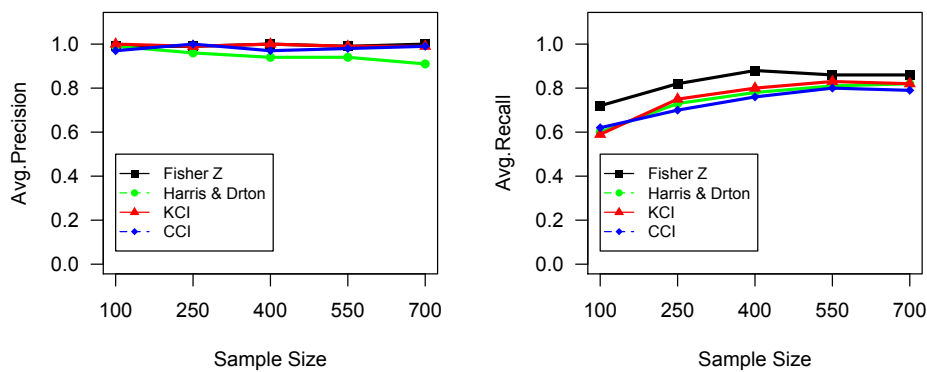

Figure 3. Adjacency precisions and recalls for the linear Gaussian simulation with setup similar to that of Zhang et al. (2011), Section 4.2.1, as a function of sample size.

The precision of all four tests is quite high, in fact almost indistinguishable per test, while recall increases substantially with sample size, with parametric Fisher Z doing the best out of all of the tests. This is somewhat discrepant with results in Zhang et al. (2011); the discrepancy can only be in the style of simulation, which for the tables above consists in i.i.d. recursive sampling. In total, KCI shows a slight edge over CCI, though all tests are pretty much on par.

Running time is more difficult to assess, since the comparison is cross-platform, and parallelization would render both KCI and CCI faster. We make all comparisons on a Macbook Air, 2 GHz Intel Core I7, using one processor. On this platform, 1000 conditional independence tests for the Fisher Z test took 0.1 seconds, for the Harris & Drton test 0.1 seconds, for the KCI test (Matlab) 1478.2 seconds, and for CCI test (Java) 17.5 seconds, for a data set simulated as above with 700 samples. The test was conditional with one conditioning variable. Thus, CCI is considerably more scalable than KCI, although the simpler Fisher Z and Harris & Drton tests are significantly more scalable than CCI. However, both KCI and CCI are amenable to parallelization. KCI can take advantage of native threading for matrix operations in Matlab to produce much faster speeds when many cores are available, and CCI can take advantage of parallelization either in the PC algorithm (through the variant PC-Stable (Colombo and Maathuis, 2013), which renders decisions about edge removals independent within each depth of execution of the PC algorithm), or within the CCI algorithm itself, which can be parallelized in the step CCI.4, above (Box 2).

---

[5] Zhang et al. simulated data using Gaussian and noise kernels; we simulate using generalized structural equation models directly.
[6] Notably, CCI uses a False Discovery Rate (FDR) step (Benjamini and Hochberg, 1995), so higher alpha values are generally tolerated.

7. Comparison on Nonlinear, Non-Gaussian Data.

One of the reasons to use a nonlinear non-Gaussian test of conditional independence is to accommodate connection functions or distributions that are not linear or Gaussian. We do the following test. Models are constructed by selecting a random DAG with 5 nodes and 5 edges, then parameterizing this as a generalized structural equation model (generalized SEM) in which the connection function of each node given its parents is as given in Table 3; there are 13 different types of connection functions; for the 14th model, connection types of types 1 though 13 are randomly selected for each variable given its parents. In each case, coefficients are drawn from U(-1, 1) and disturbance terms are distributed as U(-1, 1). For each data set, we draw 1000 i.i.d. samples according to the connection function(s). PC was then run on each data set and adjacency precisions and recalls calculated for each estimated pattern with respect to the pattern of its generating model (Figure 4) as described above for the linear, Gaussian simulation (Figure 3). An alpha value of 0.01 was used, as suggested by Zhang et al. (2011). For CCI, a polynomial basis was used, as described earlier.

Many of these results show high precision, on par with or dominating Fisher Z. Among model types with precision across all tests greater than 0.9 are 1, 2, 7, 8, and 11. However, there are a number of model types for which KCI and CCI do significantly better on precision than Fisher Z or Harris & Drton, including 3, 5, 6, 9, 10, 13, and 14. Models in which precision is not particularly good for any test are possibly 5 and 12 (though KCI gets 0.85 but with poor recall) . Notably, 12 is a model with multiplicative noise. For recall, Fisher Z gets recalls less than 0.4 for model types 3, 5, 6, 10, and 12; Harris & Drton for 6, 9, 10, and 12, KCI for just 10 and 12, and CCI for 5. Overall, KCI and CCI are roughly comparable (except for the reciprocal function case, 5, where KCI is clearly better in recall and the precisions are comparable) and for nonlinear relations both are markedly superior to the Fisher and Harris & Drton procedures.

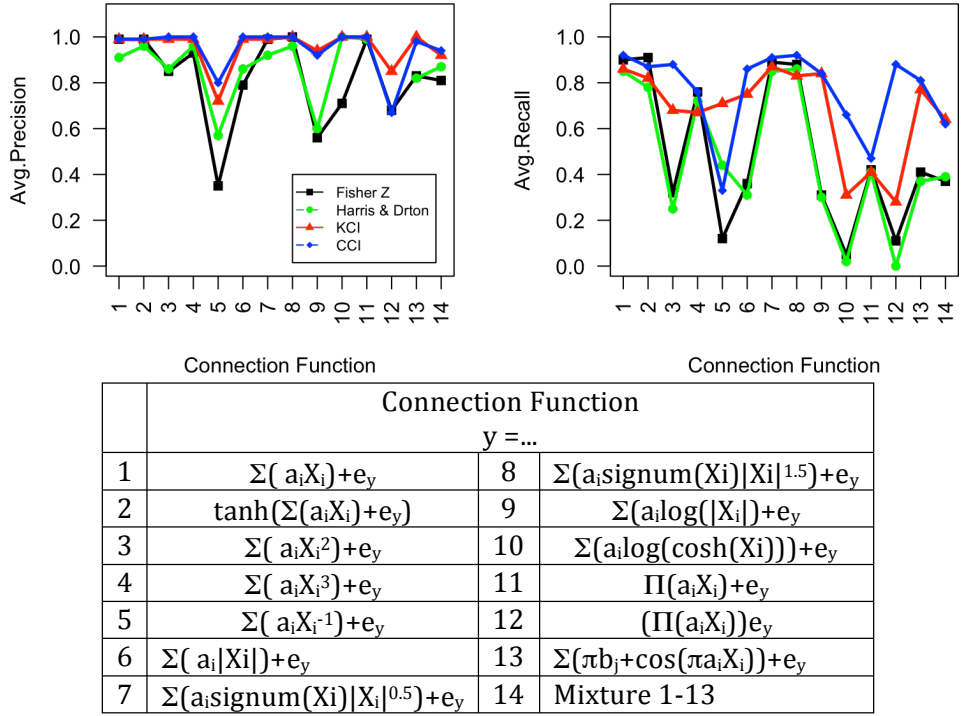

| | Connection Function | | |
|---|---|---|---|
| | y =... | | |
| 1 | $\Sigma(a_iX_i)+e_y$ | 8 | $\Sigma(a_i signum(X_i)|X_i|^{1.5})+e_y$ |
| 2 | $tanh(\Sigma(a_iX_i)+e_y)$ | 9 | $\Sigma(a_i log(|X_i|))+e_y$ |
| 3 | $\Sigma(a_iX_i^2)+e_y$ | 10 | $\Sigma(a_i log(cosh(X_i)))+e_y$ |
| 4 | $\Sigma(a_iX_i^3)+e_y$ | 11 | $\Pi(a_iX_i)+e_y$ |
| 5 | $\Sigma(a_iX_i^{-1})+e_y$ | 12 | $(\Pi(a_iX_i))e_y$ |
| 6 | $\Sigma(a_i|X_i|)+e_y$ | 13 | $\Sigma(\pi b_j+cos(\pi a_iX_i))+e_y$ |
| 7 | $\Sigma(a_i signum(X_i)|X_i|^{0.5})+e_y$ | 14 | Mixture 1-13 |

Figure 4. Average adjacency precision and recall for each of 14 model types with the specified connection functions, for random DAGs with 5 nodes and 5 edges, with coefficients drawn from U(-1, 1) and disturbance terms distributed as U(-1, 1). Sample size was 1000. Average precision for Harris & Drton for connection function 12 is undefined, average recall=0.

To illustrate the ability of PC using CCI to scale up, we choose a sparse DAG with 200 nodes and 200 edges, parameterize this using equations of the form y = $\Sigma(a_i log(cosh((X_i))+e_y$ , $\varepsilon_y \sim$ U(-1,1), $a_i \sim$ U(-1, 1), and draw 2000 i.i.d. samples. In order to parallelize the search for CCI, we use the PC-Stable algorithm (Colombo and Maathuis, 2013) instead of the PC algorithm, otherwise using the same conditions as for Figure 4. This same parallelized algorithm was used for Fisher Z and Harris & Drton. KCI was run in Matlab using the PC implementation in the Bayes Net Toolbox (BNT; Murphy, 2002), with native parallelization of matrix operations. Additionally, the first, time consuming step of Harris & Drton, calculating Kendall's Tau for each pair of variables, was parallelized. All algorithms were run on an 8 processor 3.4 GHz machine with 16G of RAM. Results are shown in Table 1.

| | Precision | Recall | Elapsed |
|---|---|---|---|
| Fisher Z | 0.21 | 0.18 | 2 s |
| Harris & Drton | 0.90 | 0.10 | 43 s |
| KCI | 0.84 | 0.48 | 14.5 h |
| CCI | 0.96 | 0.75 | 54 s |

Table 1. Adjacency precision and recall for PC-Stable (Colombo and Maathuis, 2013) using Fisher Z, Harris & Drton, KCI and CCI as independence tests on a 200 node 200 edge 2000 sample problem. See text for details.

8. Discussion

We have given a conditional independence test for the nonlinear, non-Gaussian case that is as accurate as the best available test, KCI, using Reproducing Kernel Hilbert Spaces (RKHS) and in terms of speed in a Java implementation easily exceeds that of KCI. The upshot is that nonlinear, non-Gaussian testing can be scaled up to much larger problems, both in terms of greater sample sizes and in terms of larger numbers of variables. The test can no doubt be rendered even faster by implementing it in C++ or Fortran, or any lower level language in which a Normal CDF function is available, in addition to other standard statistical functions. We are in the process of doing such implementations, in which case the test will be available for use in Matlab and in R. The test has a simpler construction than the RKHS tests, relying primarily on a theorem in Daudin (1980); this is one of the reasons its implementation in lower level languages is so straightforward. CCI does however contain a quadratic step in the estimation of residuals nonparametrically; this limits its applicability to very large data sets, say, with sample sizes greater than 10,000, with high dimension, though in these cases the test itself can easily be parallelized, yielding better performance when many processors are available.

We compare CCI with KCI on the one hand, as a representative test using the RKHS paradigm, but also to Fisher Z, a paradigmatic partial correlation test (linear, Gaussian), and the Harris & Drton test, an interesting extension of partial correlation to the Gaussian copula, and thus a partial extension into the nonlinear, non-Gaussian regime of model types. We find that on linear, Gaussian data the tests all perform on par but that on nonlinear, non-Gaussian data KCI and CCI both outperform Fisher Z and Harris & Drton, except where models use connection functions that are very close to linear, or at least have strong linear trends.

As for the form of the CCI test itself, choices can be made in some places. In the calculation of nonparametric residuals, a kernel is used. Both the width and the shape of this kernel can be adjusted. The width of the kernel needs to narrow appropriately to allow for convergence in the large sample limit, but can be generally widened or narrowed for particular types of data or sample sizes. A theoretically motivated width is used in the simulations. For shape, various positive definite shapes can be used; we use a Uniform kernel to make the convergence argument straightforward. In any case, the width of the kernel has more influence on performance than the shape of the kernel.

Another place in the algorithm where some discretion can be exercised is in the choice of the truncated basis F of functions in the procedure CCI. There is considerable flexibility on this point, since in general one needs to include just as many functions in F as are required to break the relevant symmetries that exist in distributions for variables. In Table 3, 13 connection function types are given; for the most part, F = {$x, x^2$} is sufficient, since functions either have a linear trend (captured by $f_1(x) = x$) or are symmetric (captured by $f_2(x) = x^2$), where $f_2(x)$ has a linear trend, though adding more functions into F can improve accuracy, capturing more types of variation. In addition, standardizing or otherwise scaling down data with particularly large values may be helpful, to bring them into a range where basis functions can be applied to good effect.

Acknowledgements. Research was supported by a grant from the James S. McDonnell Foundation. I thank Clark Glymour and Peter Spirtes for advice, and Ruben Sanchez for proofreading and formatting of images.